%% file: main.tex
\documentclass[a4paper]{article}

\input{packages.tex}
\usepackage[a4paper, total={6in, 8in}]{geometry}
\usepackage{csquotes}
\usepackage[backend=bibtex,style=numeric-comp,giveninits=true,doi=false,isbn=false,url=false,eprint=false,maxbibnames=99]{biblatex}

\input{macros.tex}

\graphicspath{{pictures/}}
\doublefalse

\title{{\papertitle}}
\input{header.tex}

\begin{document}
	\maketitle
	
	\begin{abstract}
		\input{parts_abstract.tex}
	\end{abstract}
	
        \newpage
	\input{parts_intro.tex}

	\input{parts_methods.tex}
	\input{parts_results.tex}
	\input{parts_discussion.tex}
    \input{parts_conclusions.tex}

	\input{acknowledgements.tex}

    \newpage
	\printbibliography

\end{document}

%% file: packages.tex
\usepackage[english]{babel}
\usepackage[utf8]{inputenc}
\usepackage[colorinlistoftodos]{todonotes}
\usepackage{amsmath}
\usepackage{amssymb}
\usepackage{amsthm} 
\usepackage{mathtools}
\usepackage{mathrsfs}
\usepackage{siunitx}
\usepackage{algpseudocode}

\usepackage{subfig}

\usepackage{tikz}

\usepackage{tabularx}
\usepackage{booktabs}
\usepackage{multirow}

\usepackage{empheq}

\usepackage{lipsum}

\usepackage[toc,page]{appendix}

\usepackage{rotating}

\usepackage{makecell}

\usepackage{lscape}
\usepackage{xcolor,colortbl}
\definecolor{lavender}{rgb}{0.9, 0.9, 0.98}

\theoremstyle{plain}
\theoremstyle{plain}
\theoremstyle{plain}
\theoremstyle{plain}
\theoremstyle{plain}
\theoremstyle{definition}
\theoremstyle{remark}

\everymath{\displaystyle}

%% file: macros.tex
\newif\ifdouble

\newcommand{\papertitle}{Branched Latent Neural \MS{Maps}}

\newcommand{\keywordOne}{Branched Latent Neural \MS{Maps}}
\newcommand{\keywordTwo}{Scientific Machine Learning}
\newcommand{\keywordThree}{Numerical Simulations}
\newcommand{\keywordFour}{Cardiac Electrophysiology}
\newcommand{\keywordFive}{Congenital Heart Disease}

\newcommand{\blackcolor}{black}
\newcommand{\MS}[1]{\textcolor{\blackcolor}{#1}}
\newcommand{\MSblue}[1]{\textcolor{\blackcolor}{#1}}

\newcommand{\modFOM}{\mathcal{M}_{\text{HF}}}
\newcommand{\modROM}{\mathcal{M}_{\text{BLNM}}}

\newcommand{\ANNState}{\mathbf{z}}
\newcommand{\ANNStateTildeObs}{\widetilde{\mathbf{z}}_\mathrm{obs}}
\newcommand{\ANNLatent}{\mathbf{z}_\mathrm{latent}}

\newcommand{\ANNRhs}{\mathcal{B L \kern0.05em N \kern-0.05em M}}
\newcommand{\NN}{\mathcal{N \kern-0.05em N}}

\newcommand{\NumANNState} {N_\mathrm{z}}
\newcommand{\NumANNStatePhysical} {N_{z_\mathrm{physical}}}

\newcommand{\NumANNWeights} {N_\mathrm{w}}

\newcommand{\ANNparam}{\mathbf{w}}
\newcommand{\ANNparamTrained}{\widehat{\ANNparam}}

\newcommand{\param}{\boldsymbol{\theta}}
\newcommand{\paramAdim}{\widetilde{\param}}
\newcommand{\paramEP}{\param_\mathrm{EP}}
\newcommand{\paramEPAdim}{\widetilde{\param}_\mathrm{EP}}
\newcommand{\paramEPAdimInit}{\widetilde{\param}_{EP}^\mathrm{init}}
\newcommand{\paramEPAdimDE}{\widetilde{\param}_\mathrm{EP}^\mathrm{DE}}
\newcommand{\paramEPAdimHF}{\widetilde{\param}_\mathrm{EP}^\mathrm{HF}}

\newcommand{\paramSpace}{\boldsymbol{\Theta}}

\newcommand{\NumParams} {N_\mathcal{P}}

\newcommand{\THB}{T_\mathrm{HB}}

\newcommand{\fZero}{{\mathbf{f}_0}}

\newcommand{\Vone}{V_\mathrm{1}}
\newcommand{\Vtwo}{V_\mathrm{2}}
\newcommand{\Vthree}{V_\mathrm{3}}
\newcommand{\Vfour}{V_\mathrm{4}}
\newcommand{\Vfive}{V_\mathrm{5}}
\newcommand{\Vsix}{V_\mathrm{6}}

\newcommand{\aVF}{aVF}
\newcommand{\one}{I}

\newcommand{\LA}{LA}
\newcommand{\RA}{RA}
\newcommand{\F}{F}

\newcommand{\Iion}{{\mathcal{I}_{\mathrm{ion}}}}
\newcommand{\Iapp}{{\mathcal{I}_{\mathrm{app}}}}
\newcommand{\DiffTens}{\boldsymbol{D}_{\mathrm{M}}}
\newcommand{\Pot}{u}
\newcommand{\PotExt}{u_\mathrm{e}}
\newcommand{\Gating}{\boldsymbol{w}}
\newcommand{\Conc}{\boldsymbol{z}}
\newcommand{\RhsGating}{\boldsymbol{H}}
\newcommand{\RhsConc}{\boldsymbol{G}}

\newcommand{\GCaL}{G_\mathrm{CaL}}
\newcommand{\GNa}{G_\mathrm{Na}}
\newcommand{\GKr}{G_\mathrm{Kr}}

\newcommand{\Da}{D_\mathrm{ani}}
\newcommand{\Di}{D_\mathrm{iso}}
\newcommand{\Dp}{D_\mathrm{purk}}
\newcommand{\tLVstim}{t_\mathrm{LV}^\mathrm{stim}}

\newcommand{\ANNStateECG}{\mathbf{z}_\mathrm{leads}}
\newcommand{\ANNStateECGAdim}{\widetilde{\mathbf{z}}_\mathrm{leads}}

\newcommand{\ANNStatePhysical}{\mathbf{z}_\mathrm{physical}}
\newcommand{\ANNStatePhysicalAdim}{\widetilde{\mathbf{z}}_\mathrm{physical}}

%% file: header.tex
\author{Matteo Salvador$^{2, 3, 4, *}$,
	    Alison Lesley Marsden$^{1, 2, 3, 4}$}

\date{\footnotesize
    $^1$ Department of Bioengineering, Stanford University, California, USA \\
	$^2$ Institute for Computational and Mathematical Engineering, Stanford University, California, USA \\
    $^3$ Cardiovascular Institute, Stanford University, California, USA \\
	$^4$ Pediatric Cardiology, Stanford University, California, USA \\
	$^*$ \textit{Corresponding author} (\texttt{msalvad@stanford.edu}) \\
    }

%% file: parts_abstract.tex
We introduce Branched Latent Neural \MS{Maps} (\MS{BLNMs}) to learn \MS{finite dimensional} input-output maps encoding complex physical processes.
A \MS{BLNM} is defined by a simple and compact feedforward partially-connected neural network that structurally disentangles inputs with different intrinsic roles, such as the time variable from model parameters of a differential equation, while transferring them into a generic field of interest.
\MS{BLNMs} leverage \MSblue{latent} outputs to enhance the learned dynamics and break the curse of dimensionality by showing excellent \MS{in-distribution} generalization properties with small training datasets and short training times on a single processor.
Indeed, their \MS{in-distribution} generalization error remains comparable regardless of the adopted discretization during the testing phase.
Moreover, the partial connections, in place of a fully-connected structure, significantly reduce the number of tunable parameters.
We show the capabilities of \MS{BLNMs} in a challenging test case involving biophysically detailed electrophysiology simulations in a biventricular cardiac model of a pediatric patient with hypoplastic left heart syndrome. The model includes a 1D Purkinje network for fast conduction and a 3D heart-torso geometry.  
Specifically, we trained \MS{BLNMs} on 150 in silico generated 12-lead electrocardiograms (ECGs) while spanning 7 model parameters, covering cell-scale, organ-level and electrical dyssynchrony.
Although the 12-lead ECGs manifest very fast dynamics with sharp gradients, after automatic hyperparameter tuning the optimal \MS{BLNM}, trained in less than 3 hours on a single CPU, retains just 7 hidden layers and 19 neurons per layer. The resulting mean square error is on the order of $10^{-4}$ on an independent test dataset comprised of 50 additional electrophysiology simulations.
In the online phase, the \MS{BLNM} allows for 5000x faster real-time simulations of cardiac electrophysiology on a single core standard computer \MS{and can be employed to solve inverse problems via global optimization in a few seconds of computational time}.
This paper provides a novel computational tool to build reliable and efficient reduced-order models for digital twinning in engineering applications.
The Julia implementation is publicly available under MIT License at \MS{\url{https://github.com/StanfordCBCL/BLNM.jl}}.

%% file: parts_intro.tex
\section{Introduction}
\label{sec:introduction}

Learning complex input-output maps behind physical processes in a reliable manner has significant implications in any field of science and engineering.
In particular, when these physical processes are described via mechanistic models, the numerical resolution of the underlying differential equations may be challenging and computationally demanding, even for a single instance of model parameters \cite{Quarteroni2016,Quarteroni2017Numerical}.

\MS{In the past few years, several methods in the field of model order reduction, partially or entirely based on Neural Networks (NNs), have been proposed to mitigate the high computational cost of physics-based solvers, with the aim of producing accurate and efficient model evaluations for many-query applications \cite{Fresca2021,Lu2021,Raonic2023,Regazzoni2023,Vlachas2022,Pegolotti2023}, which involve sensitivity analysis, parameter estimation, forward and inverse uncertainty quantification, and optimization \cite{Fleeter2020,Tran2017,Salvador2023}.
However, many intrusive \cite{Quarteroni2016} and non-intrusive \cite{Hesthaven2018} reduced-order models either fail or struggle to effectively reproduce phenomena that manifest fast and irregular dynamics while spanning an elaborate solution manifold.}
In this paper, we propose a novel computational tool which we term Branched Latent Neural \MS{Maps} (\MS{BLNMs}) to accurately and efficiently learn generic input-output relationships, even in the presence of sharp features and significant variability.
\MS{BLNMs} are based on feedforward partially-connected NNs \cite{Kang2005} to separate the contributions coming from unrelated inputs, such as space and time variables with respect to physics-based scalar parameters.
The output of \MS{BLNMs} is given by relevant scalar or vector fields of interest, as well as \MSblue{additional latent} variables, which serve the purpose of enhancing the learned dynamics.
The presence of partial connections allows for a significant reduction in the number of tunable parameters while ensuring excellent \MS{in-distribution} generalization properties during the testing phase, even on different mesh resolutions than those used during the training stage.

Several Machine Learning methods have been recently proposed to tackle cardiac electromechanics while exploiting physics-based knowledge \cite{Cicci2022,Fresca2020,Pagani2021,Tenderini2022,Regazzoni2019mor-sarcomeres, Regazzoni2022EMROM}.
In this paper, we demonstrate the performance of \MS{BLNMs} in the setting of cardiovascular modeling \cite{Fedele2023, Strocchi2020Cohort, Peirlinck2021, Pfaller2019} and congenital heart disease \cite{Marsden2015, VignonClementel2010}, where multiphysics and multiscale phenomena interact in the context of understudied pathological conditions in the field of computational cardiology.
Specifically, we consider a patient-specific heart-torso geometry of a pediatric case with hypoplastic left heart syndrome (HLHS) \cite{Feinstein2012}.  We perform biventricular-Purkinje 3D-1D electrophysiology simulations to compute in-silico 12-lead electrocardiograms (ECGs) while spanning cell-scale through tissue-level parameter variability of a biophysically detailed mathematical model of electrophysiology.
A \MS{BLNM} trained on 150 electrophysiology simulations in less than 3 hours on a single CPU, endowed with 7 hidden layers and 19 neurons per layer (2,398 tunable parameters), retains an approximation error on the order of $10^{-4}$ on 50 additional unseen 12-lead ECGs by the NN.
Moreover, it enables faster than real-time numerical simulations during the online phase, \MS{which allows one to accurately and efficiently solve inverse problems.}
\MS{Indeed, this task would be unaffordable using a biophysically detailed electrophysiology model, given the computational cost of these numerical simulations and the amount of queries that are required to solve a nonlinear optimization problem.} 
\MS{BLNMs are lightweight, compact, easy to train architectures, able to precisely capture the fast time scales of 12-lead ECGs while spanning cell-to-organ model variability.}
\MS{Moreover, they can be queried in fractions of seconds to generate new predictions.}
\MS{Overall, BLNMs} provide a novel computational tool for the generation of accurate and efficient standalone surrogate models that can be applied for digital twinning in computational science.

%% file: parts_methods.tex
\section{Methods}
\label{sec:methods}
We describe the methodological details behind \MS{BLNMs} for time-dependent processes, as well as the mathematical and numerical models adopted for the application of simulated cardiac electrophysiology in a congenital heart disease patient.

\subsection{Branched Latent Neural \MS{Maps}}
\label{sec:methods:BLNO}

\begin{figure}[t!]
    \centering
    \includegraphics[width=1.0\textwidth]{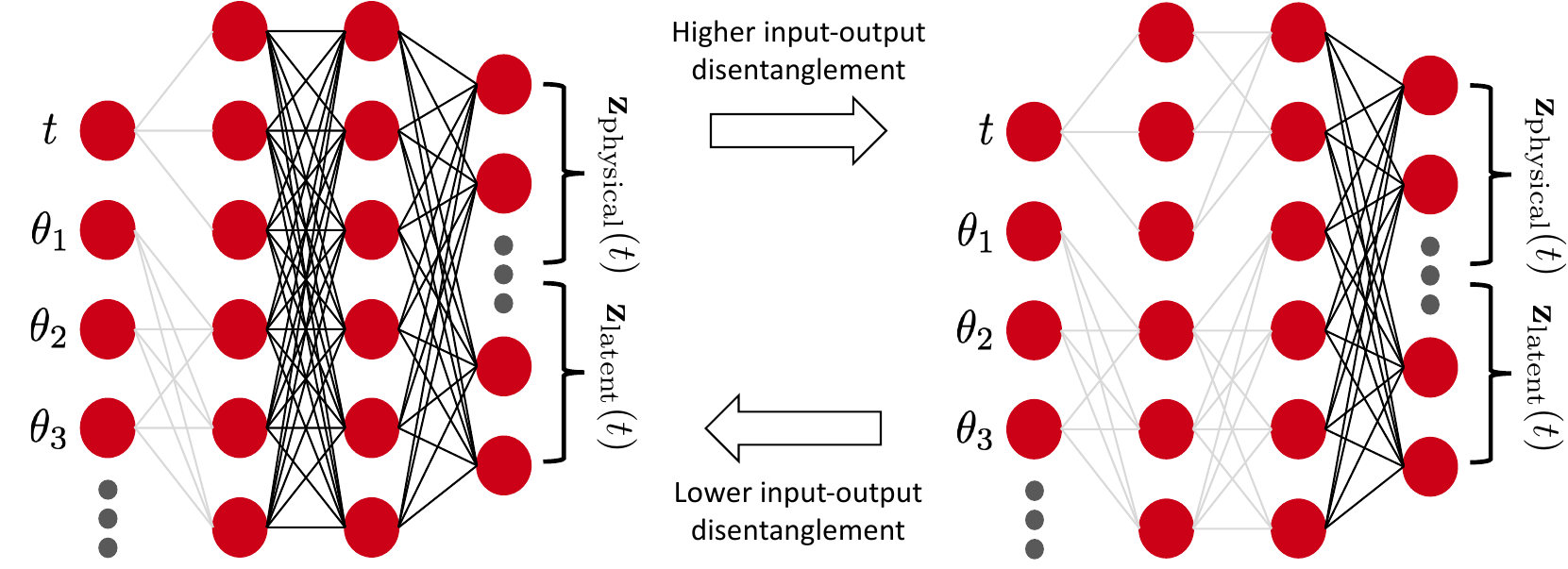}
    \caption{\MS{Sketch of Branched Latent Neural Maps with different disentanglement levels between inputs, involving time variable $t$ and model parameters $\param$, and outputs, i.e. generic fields of interest, including both physical $\ANNStatePhysical(t)$ and latent $\ANNLatent(t)$ temporal quantities. Partial connections are depicted in light grey, whereas full connections are outlined in black.}}
    \label{fig:BLNO}
\end{figure}

Given a generic high-fidelity model $\modFOM$ expressed in terms of an input-output map between model parameters and a time-dependent process, we derive a surrogate model \MS{$\modROM$} by building a feedforward partially-connected NN that explores model $\modFOM$ parametric variability while structurally separating the role of time and model parameters.
We depict the \MS{BLNM} architecture in Figure~\ref{fig:BLNO}, showing that different levels of disentanglement are allowed, ranging from the first hidden layer to the output layer.
\MS{This disentanglement enables BLNMs to generalize well over different grids during testing even if the training stage is performed on a specific finite dimensional resolution (see Section~\ref{sec:results:BLNO}).}
The surrogate model reads: 
\begin{equation} \label{eqn:BLNO}
    \ANNState(t) = \ANNRhs \left(t, \param; \ANNparam \right) \text{ for } t \in [0, T].
\end{equation}
This feedforward partially-connected NN is represented by weights and biases $\ANNparam \in \mathbb{R}^{\NumANNWeights}$, and defines a map $\ANNRhs \colon \mathbb{R}^{1 + \NumParams} \to \mathbb{R}^{\NumANNState}$ from time $t$ and model parameters $\param \in \paramSpace \subset \mathbb{R}^{\NumParams}$ to a state vector $\ANNState(t) = [\ANNStatePhysical(t), \ANNLatent(t)]^T$.
Indeed, the state vector $\ANNState(t) \in \mathbb{R}^{\NumANNState}$ contains $\ANNStatePhysical(t)$ physical fields of interest, as well as \MSblue{$\ANNLatent(t)$} latent temporal variables without a direct physical representation, that enhance the learned dynamics of the \MS{BLNM}.
\MSblue{These non-dimensional latent variables $\ANNLatent(t)$ are not accounted for in the loss function during the training stage, as in neural differential equations \cite{Dupont2019, regazzoni2019modellearning, Regazzoni2023}, but enrich the generalization of BLNMs while mapping the whole solution manifold, by selectively and properly acting in areas with steep gradients.}
During the optimization process of the NN tunable parameters, we minimize the Mean Square Error (MSE), that is:
\begin{equation} \label{eqn:loss_LNO}
\begin{split}
\mathcal{L}(\ANNStatePhysicalAdim(t), \ANNStateTildeObs(t); \ANNparamTrained) = \underset{\ANNparamTrained}{\arg\min} & \left[ || \ANNStatePhysicalAdim(t) - \ANNStateTildeObs(t) ||_{\text{L}^2(0, T)}^2 \right], \\
\end{split}
\end{equation}
where $\ANNStatePhysicalAdim(t) \in [-1, 1]^{\NumANNStatePhysical}$ and $\ANNStateTildeObs(t) \in [-1, 1]^{\NumANNStatePhysical}$ represent model \MS{$\modROM$} outputs and observations in non-dimensional form.
Time $\widetilde{t} \in [0, 1]$ and model parameters $\paramAdim \in [-1, 1]^{\NumParams}$ are also normalized during the training phase of model \MS{$\modROM$}.

\subsection{Cardiac electrophysiology}
\label{sec:methods:EP}

\begin{figure}[t!]
    \centering
    \includegraphics[width=1.0\textwidth]{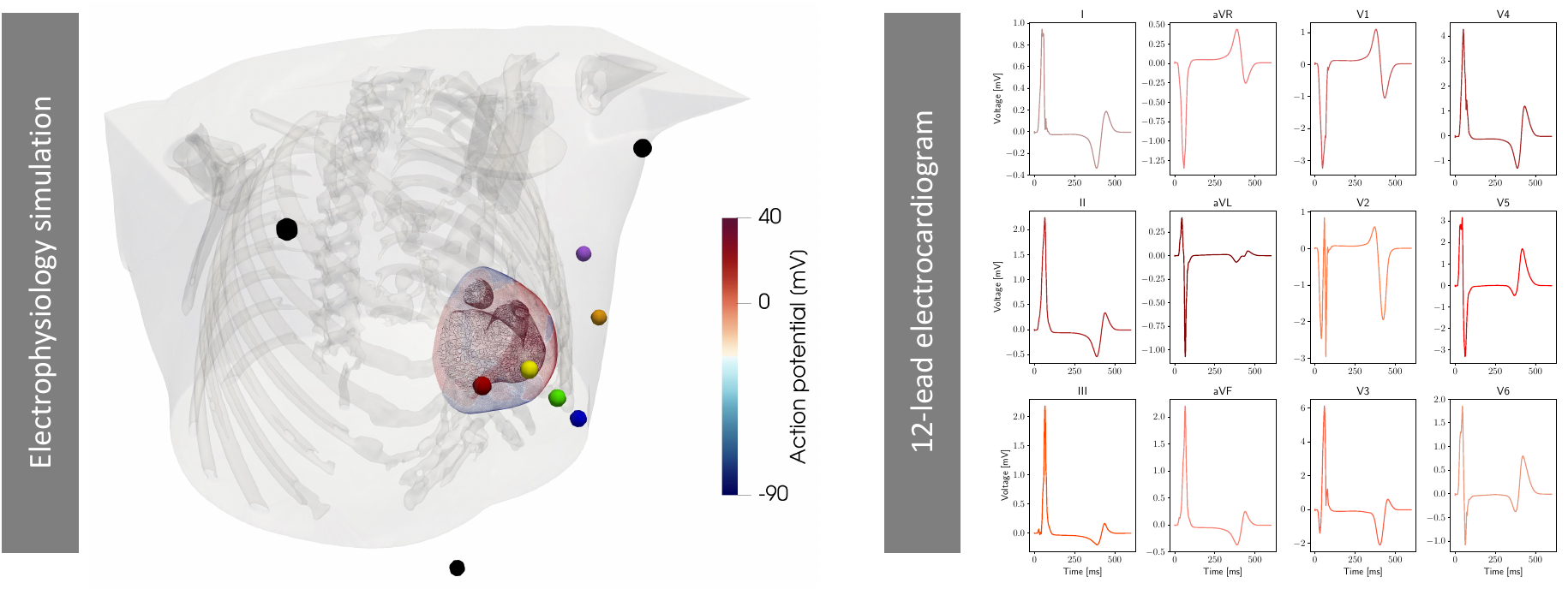}
    \caption{Example of heart-torso electrophysiology simulation in the patient-specific cardiac model and corresponding 12-lead simulated ECGs.}
    \label{fig:EP}
\end{figure}

We reconstruct a heart-torso model of a 7 year old female pediatric patient with HLHS from computerized tomography (CT) images.
\MS{Images and associated clinical data were obtained under an IRB-approved protocol at Stanford University.}
In Figure~\ref{fig:EP} we show an example of an electrophysiology simulation and in silico derived 12-lead ECGs on this patient-specific geometry.

\subsubsection{Mathematical model}
\label{sec:methods:EP:model}

We model cardiac electrophysiology in the heart-Purkinje system by considering the biophysically detailed monodomain equation \cite{Quarteroni2019,collifranzone2014book} coupled with the ten Tusscher-Panfilov ionic model \cite{TTP06}, represented here in compact form:
\begin{eqnarray} \label{eqn:monodomain}
\left\{\begin{array}{ll}
\displaystyle
\frac{\partial \Pot}{\partial t}+\Iion(\Pot,\Gating,\Conc)
-\nabla\cdot(\DiffTens\nabla \Pot)=\Iapp({\bf x},t) & \mbox{ in }\Omega\times(0,T],\\[2mm]
(\DiffTens\nabla \Pot)\cdot {\bf n}=0  & \mbox{ on }\partial\Omega\times(0,T],\\[2mm]
\displaystyle
\frac{d\Gating}{dt}=\RhsGating(\Pot,\Gating,\Conc)
& \mbox{ in }\Omega\times(0,T],\\[2mm]
\displaystyle
\frac{d\Conc}{dt}=\RhsConc(\Pot,\Gating,\Conc)
& \mbox{ in }\Omega\times(0,T],\\[2mm]
\Pot({\bf x},0)=\Pot_0({\bf x}),\
\Gating({\bf x},0)=\Gating_0({\bf x}),\
\Conc({\bf x},0)=\Conc_0({\bf x}) &  \mbox{ in }\Omega.
\end{array}\right.
\end{eqnarray}
In the following, we denote Equation~\eqref{eqn:monodomain} as model $\modFOM$. $T = \THB = 600$ ms corresponds to the final simulation time, given by a single heartbeat. The computational domain $\Omega = \Omega_\mathrm{purk} \cup \Omega_\mathrm{myo}$ is represented by the one-way coupled 1D Purkinje network and 3D biventricular patient-specific geometry.

Transmembrane potential $\Pot$ describes the propagation of the electric signal at the Purkinje and myocardial level, vector $\Gating=(w_1,\ldots,w_M)$ defines the probability density
functions of $M=12$ gating variables, which represent the fraction of open channels across the membrane of a single cardiomyocyte, and vector $\Conc
=(z_1,\ldots, z_P)$ introduces the concentrations of $P=6$ relevant ionic species. Among them, sodium $Na^{+}$, intracellular calcium $Ca^{2+}$ and potassium $K^{+}$ play an important role in the physiological processes \cite{Bartos2015} dictating heart rhythmicity or sarcomere contractility, and are generally targeted by pharmaceutical therapies \cite{Brennan2009}. 
Right hand sides $\RhsGating(\Pot,\Gating,\Conc)$ and $\RhsConc(\Pot,\Gating,\Conc)$, which describe the dynamics of the gating variables and ionic concentrations respectively, along with ionic current $\Iion(\Pot,\Gating,\Conc)$, derive from the mathematical formulation of the ten Tusscher-Panfilov ionic model \cite{TTP06}.
The action potential is triggered in the left and right bundle branches by an external applied current $\Iapp({\bf x},t)$.

The diffusion tensor is expressed as $\DiffTens = \Di \mathbf{I} + \Da \fZero \otimes \fZero$ in $\Omega_\mathrm{myo}$ and $\DiffTens = \Dp \mathbf{I}$ in $\Omega_\mathrm{purk}$, where $\fZero$ expresses the biventricular fiber field \cite{Piersanti2021}. $\Da, \Di, \Dp \in \mathbb{R}^+$ represent the anisotropic, isotropic and Purkinje conductivities, respectively. 

We impose the condition of an electrically isolated domain by prescribing homogeneous Neumann boundary conditions $\partial\Omega$, where $\mathbf{n}$ is the outward unit normal vector to the boundary.


The ECG signals $\PotExt$ are computed in each lead location ${\bf x}_\mathrm{e}$ following \cite{Costabal2018}:
\begin{equation} \label{eqn:ECGs}
\PotExt({\bf x}_\mathrm{e}) = - \int_{\Omega} \nabla \Pot \cdot \nabla \dfrac{1}{|| {\bf x} - {\bf x}_\mathrm{e}||} dV,
\end{equation}
where $e = \{ V_1, V_2, V_3, V_4, V_5, V_6 \}$ and $e = \{ LA, RA, F \}$ define 6 precordial leads and 3 limb leads located on the pediatric patient-specific torso model, respectively.
From this information, we retrieve 3 bipolar limb leads as:
\begin{equation} \label{eqn:bipolarlimb}
I   = LA - RA \;\;\;\;
II  = F - RA \;\;\;\;
III = F - LA, \;\;\;\;
\end{equation}
and 3 augmented limb leads as:
\begin{equation} \label{eqn:augmentedlimb}
aVL  = (I - III) / 2 \;\;\;\;
aVR  = -(I + II) / 2 \;\;\;\;
aVF  = (II + III) / 2.
\end{equation}
The set $ECG = \{ V_1, V_2, V_3, V_4, V_5, V_6, I, II, III, aVL, aVR, aVF \}$ defines a 12-lead ECG, which is a comprehensive representation of the electrical activity in the heart \cite{collifranzone2014book}.

In Table~\ref{tab:parameterspace} we report descriptions, ranges and units for the 7 model parameters that we explore via latin hypercube sampling to generate the dataset of 200 electrophysiology simulations.

\begin{table}[t!]
    \begin{center}
        \hspace*{-0.75cm}
        \begin{tabular}{l l l l}
            \toprule
            Parameter & Description & Range & Units \\
            \midrule
            $\GCaL$    & Maximal $Ca^{2+}$ current conductance & [1.99e-5, 7.96e-5]         & $\mathrm{cm}$ $\mathrm{ms}^{-1}$ $\mu \mathrm{F}^{-1}$ \\
            $\GNa$     & Maximal $Na^{+}$ current conductance  & [7.42, 29.68]              & $\mathrm{nS}$ $\mathrm{pF}^{-1}$ \\
            $\GKr$     & Maximal rapid delayed rectifier current conductance & [0.08, 0.31] & $\mathrm{nS}$ $\mathrm{pF}^{-1}$ \\
            $\Da$      & Anisotropic conductivity & [0.008298, 0.033192]                    & $\mathrm{mm}^{2}$ $\mathrm{ms}^{-1}$ \\
            $\Di$      & Isotropic conductivity & [0.002766, 0.011064]                      & $\mathrm{mm}^{2}$ $\mathrm{ms}^{-1}$ \\
            $\Dp$      & Purkinje conductivity & [1.0, 3.5]                                 & $\mathrm{mm}^{2}$ $\mathrm{ms}^{-1}$ \\
            $\tLVstim$ & Purkinje left bundle stimulation time & [0, 100]                   & $\mathrm{ms}$ \\
            \bottomrule
        \end{tabular}
        \caption{Parameter space sampled via latin hypercube for the numerical simulations performed with model $\modFOM$.}
        \label{tab:parameterspace}
    \end{center}
\end{table}

\subsubsection{Numerical discretization}
\label{sec:methods:EP:discretization}

We perform space discretization of model $\modFOM$ using $\mathbb{P}_1$ Finite Elements.
The biventricular tetrahedral mesh is comprised of 933,916 cells and 158,277 DOFs.
The average mesh size is $h = 1$ mm.
We generate the Purkinje network for both ventricles using the fractal tree and projection algorithm proposed in \cite{Costabal2016}.
We initiate the left and right bundles from the endocardial locations near the atrioventricular node. The left bundle consists of 14,820 elements (14,821 DOFs), whereas the right bundle has 67,456 elements (67,457 DOFs).
Following the approach adopted in \cite{Tikenogullari2023}, we use non-Gaussian quadrature rules to recover convergent conduction velocities in the cardiac tissue \cite{Pezzuto2016, Woodworth2022}.
We consider a transmural variation of ionic conductances to differentiate epicardial, myocardial and endocardial properties according to \cite{TTP06}. 
For time discretization, we first update the variables of the ionic model and then the transmembrane potential by employing an Implicit-Explicit numerical scheme \cite{Regazzoni2022,Piersanti2022,Fedele2023}.
Specifically, in the monodomain equation, the diffusion term is treated implicitly and the ionic term is treated explicitly. Moreover, the ionic current is discretized by means of the Ionic Current Interpolation scheme \cite{Krishnamoorthi2013}.
\MS{We employ a fixed time step $\Delta t=0.1$ ms.}
The fiber architecture is prescribed according to the Bayer-Blake-Plank-Trayanova algorithm with $\alpha_\mathrm{epi}$ = $-60^\circ$, $\alpha_\mathrm{endo}$ = $60^\circ$, $\beta_\mathrm{epi}$ = $20^\circ$ and $\beta_\mathrm{endo}$ = $-20^\circ$ \cite{Bayer2012}.

\subsubsection{Integration with Branched Latent Neural \MS{Maps}}
In the present application, \MS{BLNMs} are used to learn in silico ECGs while spanning relevant parameters of the monodomain equation and ten Tusscher-Panfilov ionic model.
The vector $\param$ corresponds to the 7 model parameters $\paramEP = [\GCaL, \GNa, \GKr, \Da, \Di, \Dp, \tLVstim]^T$ reported in Table~\ref{tab:parameterspace}.
The vector of physical variables $\ANNStatePhysical(t)$ contains the $\ANNStateECG(t)$ precordial and limb leads recordings, that is $\ANNStateECG(t) = [\Vone(t), \Vtwo(t), \Vthree(t), \Vfour(t), \Vfive(t), \Vsix(t), \LA(t), \RA(t), \F(t)]^T$.
We note that these recordings are considered in their non-dimensional form $\ANNStateECGAdim(t) \in [-1, 1]^{\NumANNStatePhysical}$ during the training and testing phases.
The same holds for time $\widetilde{t} \in [0, 1]$ and model parameters $\paramEPAdim \in [-1, 1]^{\NumParams}$.

\subsection{\MS{Parameter estimation}}
\label{sec:methods:PE}

\MS{We employ model $\modROM$ in the setting of inverse problems.}
\MS{Specifically, we perform parameter calibration for $\paramEPAdim \in [-1, 1]^{\NumParams}$ to match BLNMs physical outputs $\ANNStatePhysicalAdim(t)$ to observations $\ANNStateTildeObs(t)$ coming from model $\modFOM$ by minimizing the MSE, all in non-dimensional form, that is:}
\begin{equation} \label{eqn:loss_LNO}
\MS{\mathcal{L}(\ANNStatePhysicalAdim(t), \ANNStateTildeObs(t)) = || \ANNStatePhysicalAdim(t) - \ANNStateTildeObs(t) ||_{\text{L}^2(0, T)}^2.} \\
\end{equation}

\MS{We randomly initialize $\paramEPAdimInit$ in the $[-1, 1]^{\NumParams}$ hypercube and we aim to recover model $\modFOM$ parameters $\paramEPAdimHF$.}
\MS{We run a single trial of an Adaptive Differential Evolution algorithm for global optimization \cite{Zhang2009}, which leads to a set of tuned model parameters $\paramEPAdimDE$ via BLNMs.}

\subsection{Software and hardware}
\label{sec:methods:software}
We employ 3D slicer \cite{Fedorov2012} for the manual segmentation of the medical images in order to reconstruct the heart-torso geometry.
Meshing of this anatomic model is carried out using the TetGen library available in the SimVascular open-source software \cite{Updegrove2017}.
All electrophysiology simulations with model $\modFOM$ are performed using \texttt{svFSIplus} \cite{Zhu2022}, a C++ high-performance computing multiphysics and multiscale finite element solver for cardiac and cardiovascular modeling, on 336 cores of the Stanford Research Computing Center.
This solver is part of the SimVascular software suite for patient-specific cardiovascular modeling \cite{Updegrove2017}.
We train model \MS{$\modROM$} by using \MS{\texttt{BLNM.jl}} \cite{Innes2018,Rackauckas2017,Rackauckas2020}, a new, in-house, Julia library for Scientific Machine Learning which is made publicly available under MIT License at \MS{\url{https://github.com/StanfordCBCL/BLNM.jl}} with this work.
This public repository also contains the dataset encompassing all the electrophysiology simulations used for the training and testing phases.

%% file: parts_results.tex
\section{Results}
\label{sec:results}

We report numerical results related to the electrophysiology simulations that were run to generate the training, validation and testing datasets for \MS{BLNMs}.
Then, we explain the technical details behind the automatic \MS{BLNM} hyperparameter tuning method and show the properties and results associated with model \MS{$\modROM$}.

\subsection{Electrophysiology simulations}
\label{sec:results:EP}

\begin{figure}[t!]
    \centering
    \includegraphics[width=0.8\textwidth]{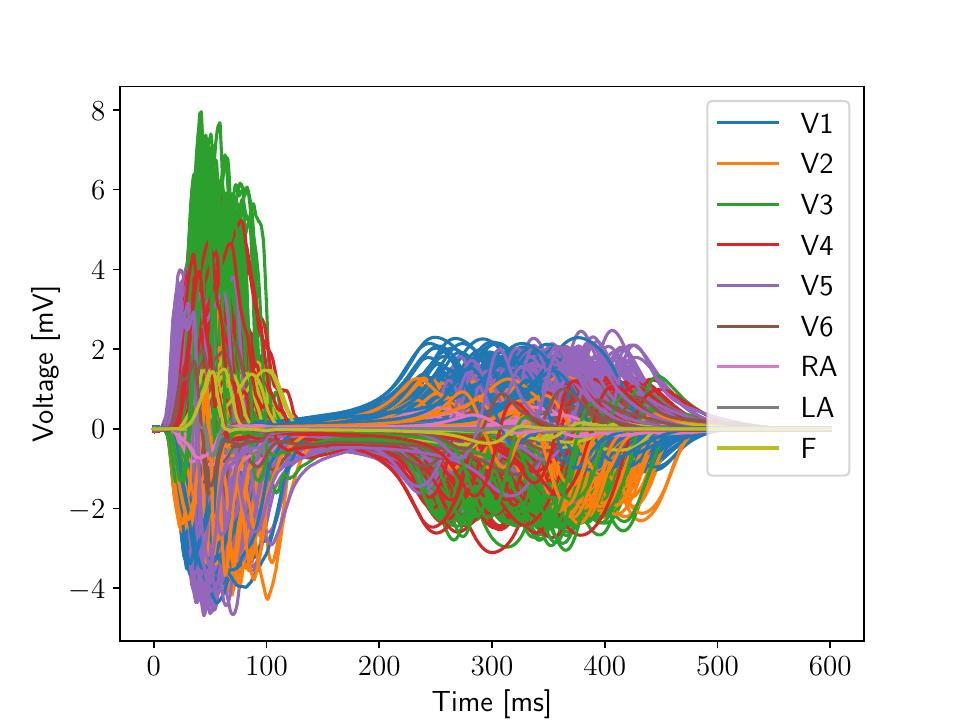}
    \caption{Full dataset containing 200 in silico precordial and limb leads recordings.}
    \label{fig:leads}
\end{figure}

\begin{figure}[t!]
    \centering
    \includegraphics[width=1.0\textwidth]{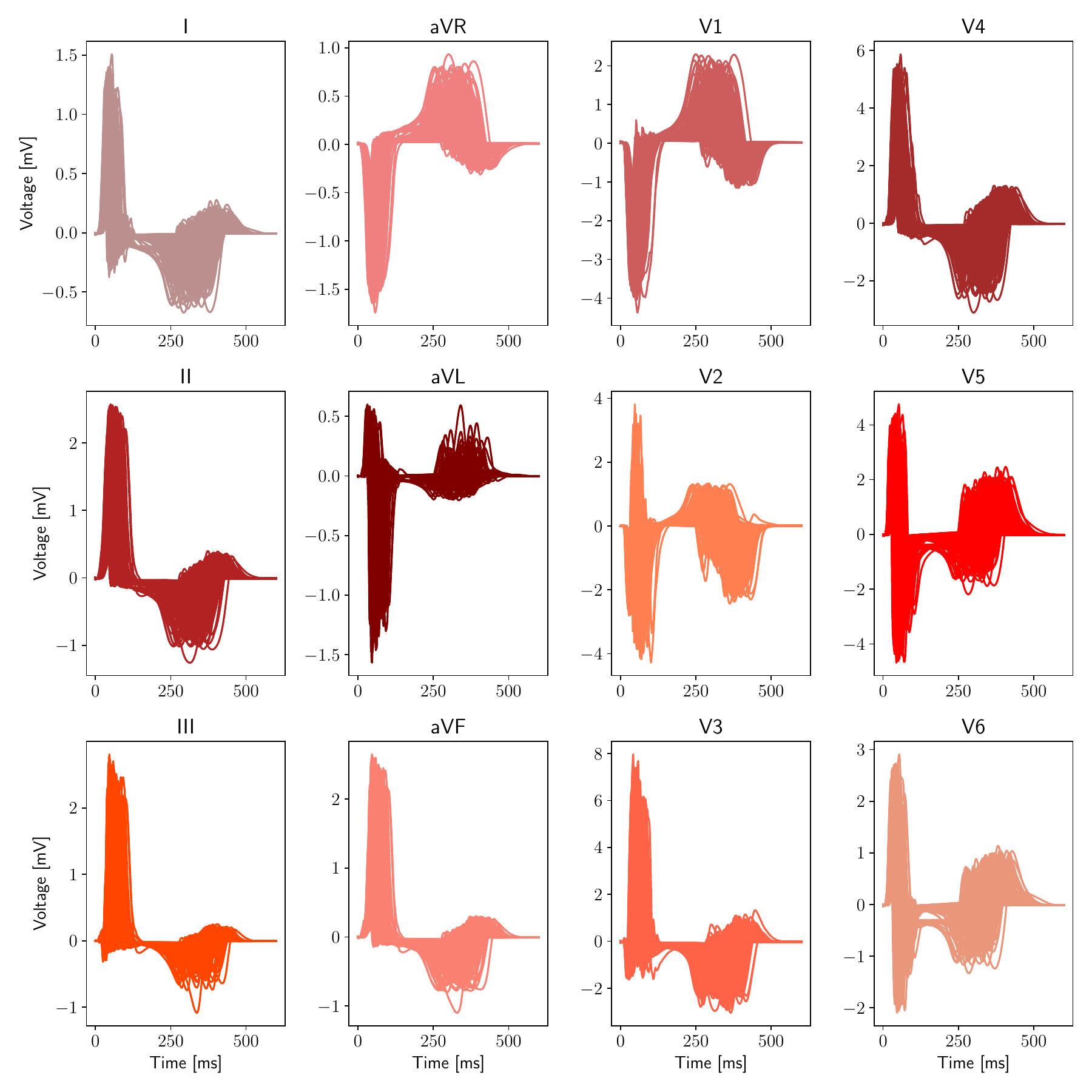}
    \caption{Full dataset containing 200 simulated 12-lead ECGs.}
    \label{fig:ECGs}
\end{figure}

\begin{figure}[t!]
    \centering
    \includegraphics[width=1.0\textwidth]{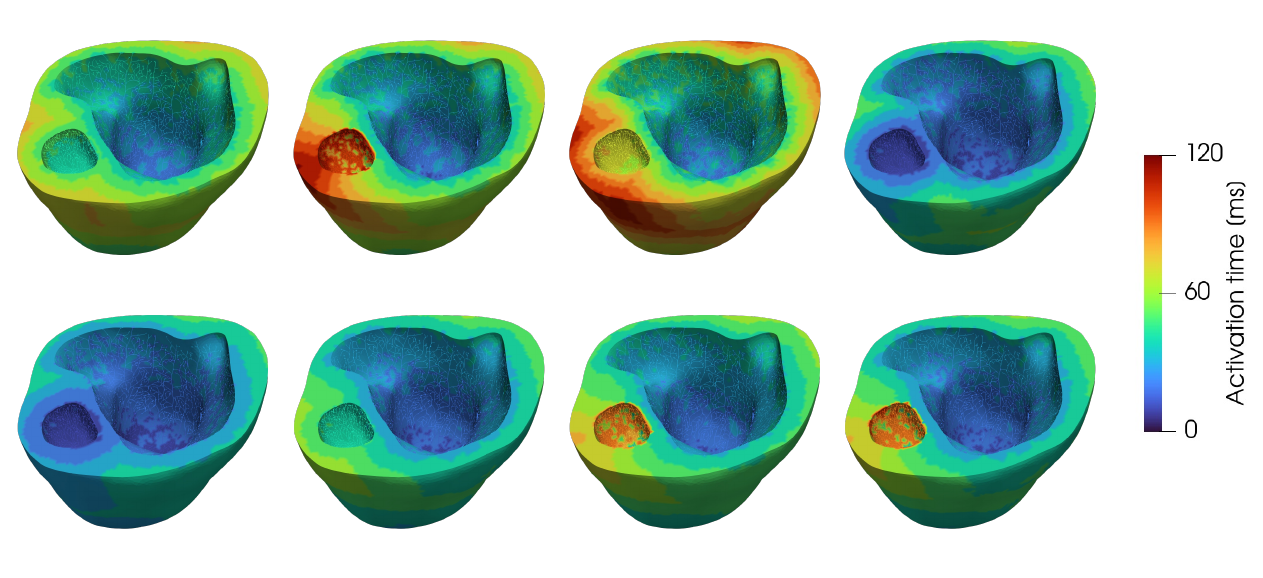}
    \caption{Simulated activation times in 8 different electrophysiology simulations that are randomly extracted from the full dataset.}
    \label{fig:AT}
\end{figure}

We ran 200 numerical simulations on the patient-specific heart-torso model (see Figure~\ref{fig:EP}) and collected the corresponding simulated 12-lead ECGs.
In Figure~\ref{fig:leads} we depict the 200 precordial and limb lead sources that are employed for training, validation and testing of the \MS{BLNMs}.
In Figure~\ref{fig:ECGs} we show the corresponding 12-lead ECGs, where the limb leads are algebraically manipulated according to Equations~\eqref{eqn:bipolarlimb} and~\eqref{eqn:augmentedlimb}.
In Figure~\ref{fig:AT} we report a representative output from the 3D electrophysiology simulation, namely activation times for 8 random samples from the whole dataset.

We notice that by exploring relevant parameters affecting cardiac function at the cell-level and organ-scale, we are able to generate a broad set of plausible 12-lead ECGs and different patterns in the activation sequence for this pediatric patient.
In particular, we remark that the simulated 12-lead ECGs produce sharp gradients during the QRS complex (ventricular depolarization) and T wave propagation (ventricular repolarization).
Moreover, they manifest high variability among different instances of the model parameters.

\subsection{Hyperparameter tuning}
\label{sec:results:tuning}

\begin{table}[h!]
    \hspace*{0.0cm}
    \begin{tabular}{c cccc c}
        \toprule
        \multirow{2}{*}{\MS{BLNM}} & \multicolumn{4}{c}{Hyperparameters} & Trainable parameters \\
               & layers & neurons & number of states & disentanglement level & \# parameters \\
        \midrule
        tuning & $\{1 \; ... \; 8\}$ & $\{10 \; ... \; 30\}$ & $\{9 \; ... \; 12\}$ & $\{1 \; ... \; N_\mathrm{layers} \}$ \\
        final    & 7              & 19                   & 10 & 2 & 2,398 \\
        \bottomrule
    \end{tabular}
    \caption{Hyperparameters ranges and selected values for the final training stage.}
    \label{tab:hyperparameters}  
\end{table}

We perform hyperparameter tuning by employing $K$-fold ($K = 5$) cross validation over 150 electrophysiology simulations.
We consider a hypercube as a search space for the number of layers, number of neurons, number of states $\NumANNState$ and disentanglement level in the \MS{BLNM} structure.
Given the limited dimension of the search space, we employ 50 instances of latin hypercube sampling and select the configuration providing the lowest MSE.
The Julia implementation is based on \texttt{Hyperopt.jl} \cite{Bagge2018}, a package to perform parallel hyperparameter optimization.
The different NNs associated with each $K$-fold are simultaneously trained via Message Passing Interface (MPI) on 5 physical cores of a standard workstation computer.
We also exploit Hyper-Threading over 7 additional virtual cores with Open Multi-Processing (OpenMP) to speed-up computations.
For each configuration of hyperparameters, we sample the dataset with a fixed time step of $\Delta t = 5.0$ ms and we perform 10,000 iterations of the second-order BFGS optimizer \cite{Goodfellow2016}.
In Table~\ref{tab:hyperparameters} we report the initial hyperparameter ranges for tuning and the final optimized values.
In Table~\ref{tab:computationaltimes} we detail the computational times and resources that we employ to generate electrophysiology simulations and to train NNs.
Generating the dataset of biophysically detailed and anatomically accurate electrophysiology simulations and reaching the final \MS{BLNM} configuration require less than 2 days of computation time.
\MS{Each electrophysiology simulation runs in approximately 10 minutes but requires hundreds of cores to achieve this performance.}
\MS{On the other hand}, training a single NN defining a \MS{BLNM} requires 10 minutes to 3 hours on a single CPU depending to the specific architecture. 

\begin{table}[t!]
    \begin{center}
        \hspace*{-1.0cm}
        \begin{tabular}{l c cc cc c}
            \toprule
            Task & Computational resources & Execution time \\
            \midrule
            Dataset generation using $\modFOM$ (200 simulations)      & 336 cores & 1 day \\
            \MS{$\modROM$} hyperparameters tuning (50 confs, 10,000 iters) & 5 cores   & 20 hours \\
            \MS{$\modROM$} final training (50,000 iters)                   & 1 core    & 2 hours and 30 minutes \\
            \bottomrule
        \end{tabular}
        \caption{Summary of the computational times and resources to generate the electrophysiology simulations with model $\modFOM$ and to train model \MS{$\modROM$}. We always tune NN parameters with the BFGS optimizer, by employing either 5 cores or serial execution on an Intel(R) Core(TM) i7-8700 3.20GHz CPU. We sample in silico 12-lead ECGs with a fixed time step $\Delta t = 5.0$ ms.}
        \label{tab:computationaltimes}
    \end{center}
\end{table}

\subsection{Branched Latent Neural \MS{Maps}}
\label{sec:results:BLNO}

\begin{table}[t!]
    \begin{center}
        \begin{tabular}{c c c c}
            \toprule
            Number of simulations & Training loss (MSE) & Testing loss (MSE) & Training time \\
            \midrule
            50  & 0.000599 & 0.018932 & 50 minutes \\
            100 & 0.000293 & 0.000589 & 1 hour and 45 minutes \\
            150 & 0.000340 & 0.000454 & 2 hours and 30 minutes \\
            \bottomrule
        \end{tabular}
        \caption{MSE and computational times associated with different training dataset with increasing complexity for the optimal NN architecture (7 layers, 19 neurons per layer, 10 states, 2 disentanglement level). We use a fixed time step $\Delta t = 5.0$ ms. We employ 1 core of a standard computer endowed with an Intel(R) Core(TM) i7-8700 3.20GHz CPU.}
        \label{tab:trainingsize}
    \end{center}
\end{table}

\begin{table}[t!]
    \begin{center}
        \begin{tabular}{c c c c}
            \toprule
            Training time step [ms] & Testing time step [ms] & Training loss (MSE) & Testing loss (MSE) \\
            \midrule
            5.0 & 0.1  & 0.000348 & 0.000459 \\
            5.0 & 1.0  & 0.000348 & 0.000458 \\
            5.0 & 5.0  & 0.000340 & 0.000454 \\
            5.0 & 10.0 & 0.000337 & 0.000452 \\
            5.0 & 20.0 & 0.000333 & 0.000445 \\
            \bottomrule
        \end{tabular}
        \caption{Testing errors associated with different sampling time steps on in silico 12-lead ECGs for the optimal NN architecture (7 layers, 19 neurons per layer, 10 states, 2 disentanglement level). We consider 50,000 BFGS iterations and 150 electrophysiology simulations for the training stage.}
        \label{tab:testingtimestep}
    \end{center}
\end{table}

\begin{table}[t!]
    \begin{center}
        \begin{tabular}{c c c c}
            \toprule
            Number of states & Training loss (MSE) & Testing loss (MSE) \\
            \midrule
            9  & 0.000758 & 0.097660 \\
            10 & 0.000340 & 0.000454 \\
            11 & 0.000358 & 0.000754 \\
            \bottomrule
        \end{tabular}
        \caption{Training and testing errors associated with different number of states on in silico 12-lead ECGs for the optimal NN architecture (7 layers, 19 neurons per layer, 2 disentanglement level). We consider 50,000 BFGS iterations and 150 electrophysiology simulations for the training stage, with $\Delta t = 5.0$ ms.}
        \label{tab:testinglatent}
    \end{center}
\end{table}

\begin{figure}[t!]
    \centering
    \includegraphics[width=1.0\textwidth]{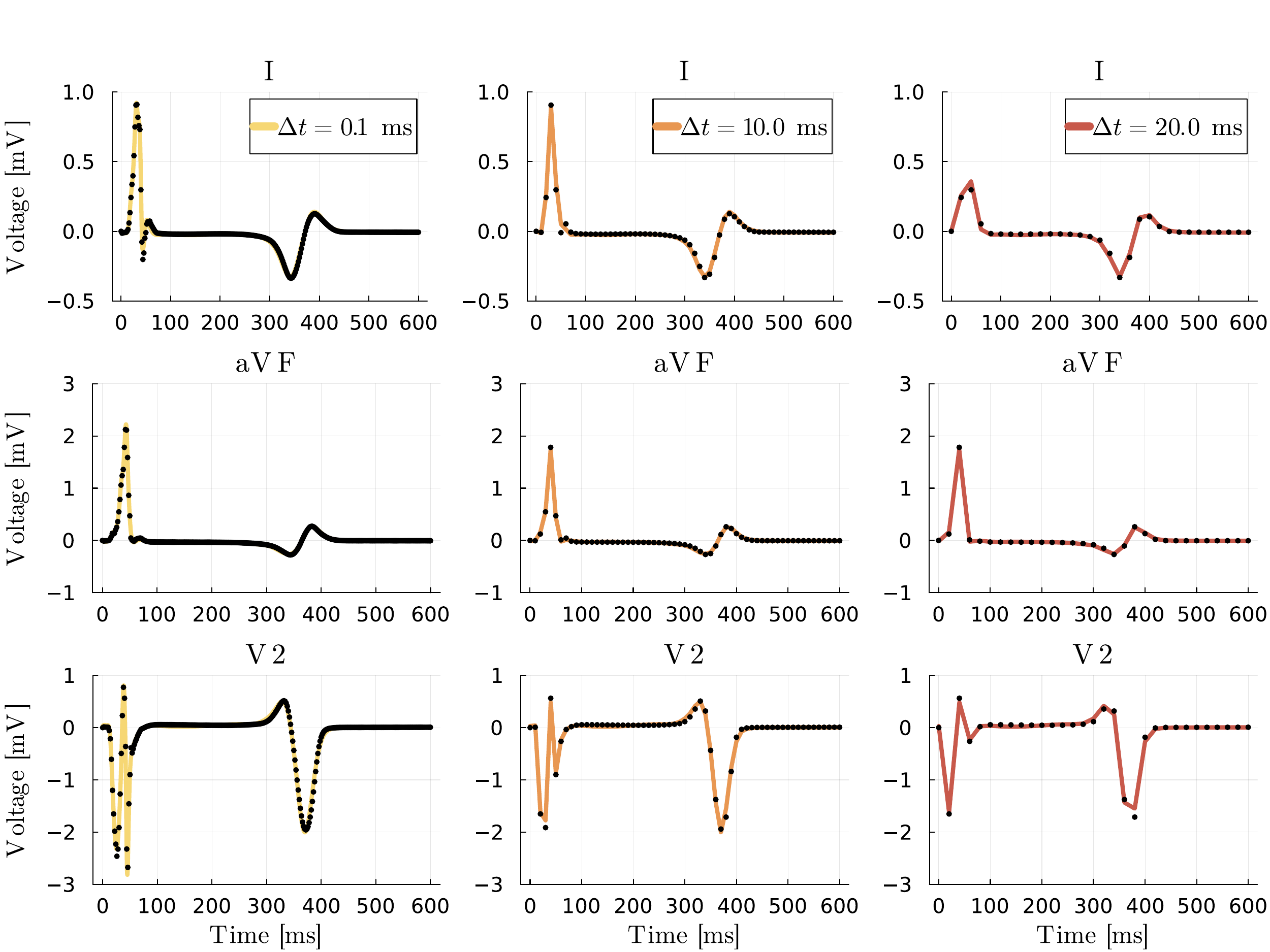}
    \caption{\MS{BLNM predictions (solid) and ground truth (points) for 1 randomly selected 12-lead ECGs in the testing set. Different colors represent different testing time steps, namely $0.1$, $10.0$ and $20.0$ ms (left to right), respectively. We show the time evolution of three relevant leads, i.e. $\one$, $\aVF$ and $\Vtwo$. We employ $\Delta t = 5.0$ ms for training.}}
    \label{fig:BLNOtestdt}
\end{figure}

\begin{figure}[t!]
    \centering
    \includegraphics[width=1.0\textwidth]{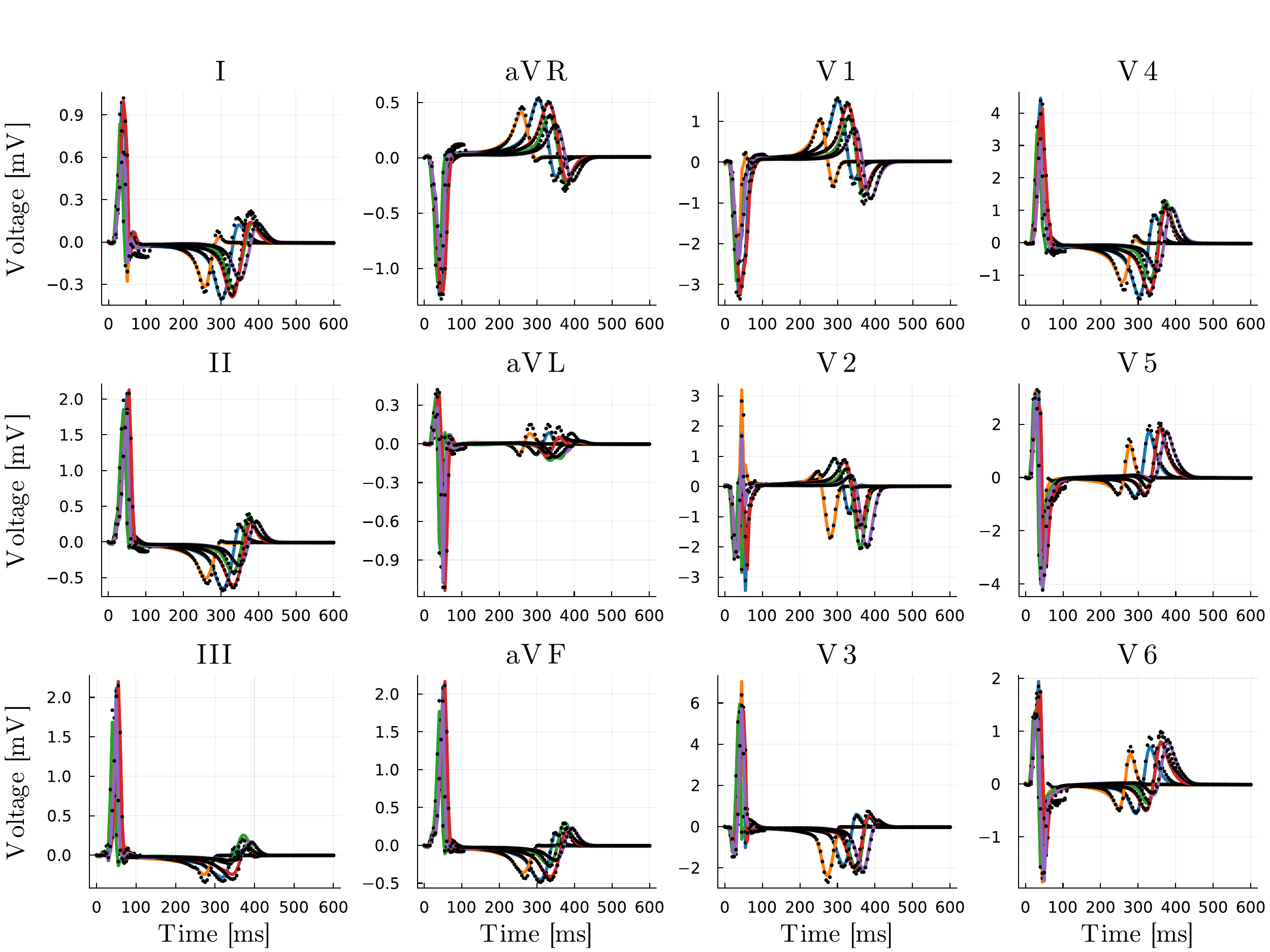}
    \caption{\MS{BLNM} predictions (solid) and ground truth (points) for 5 randomly selected 12-lead ECGs in the testing set.}
    \label{fig:BLNOtestphysical}
\end{figure}

\begin{figure}[t!]
    \centering
    \includegraphics[width=0.7\textwidth]{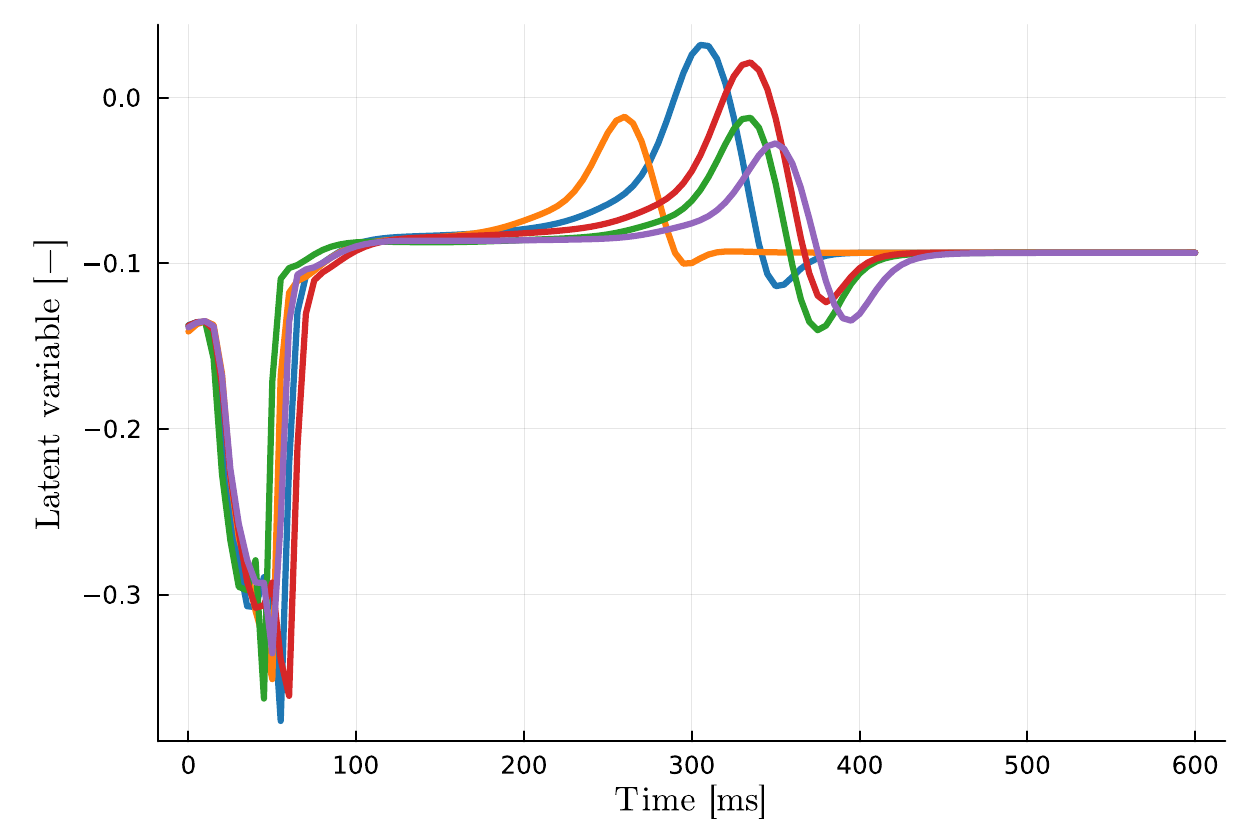}
    \caption{Time evolution of \MS{BLNM} latent variable for 5 randomly selected 12-lead ECGs in the testing set.}
    \label{fig:BLNOtestlatent}
\end{figure}

\begin{figure}[t!]
    \centering
    \includegraphics[width=0.7\textwidth]{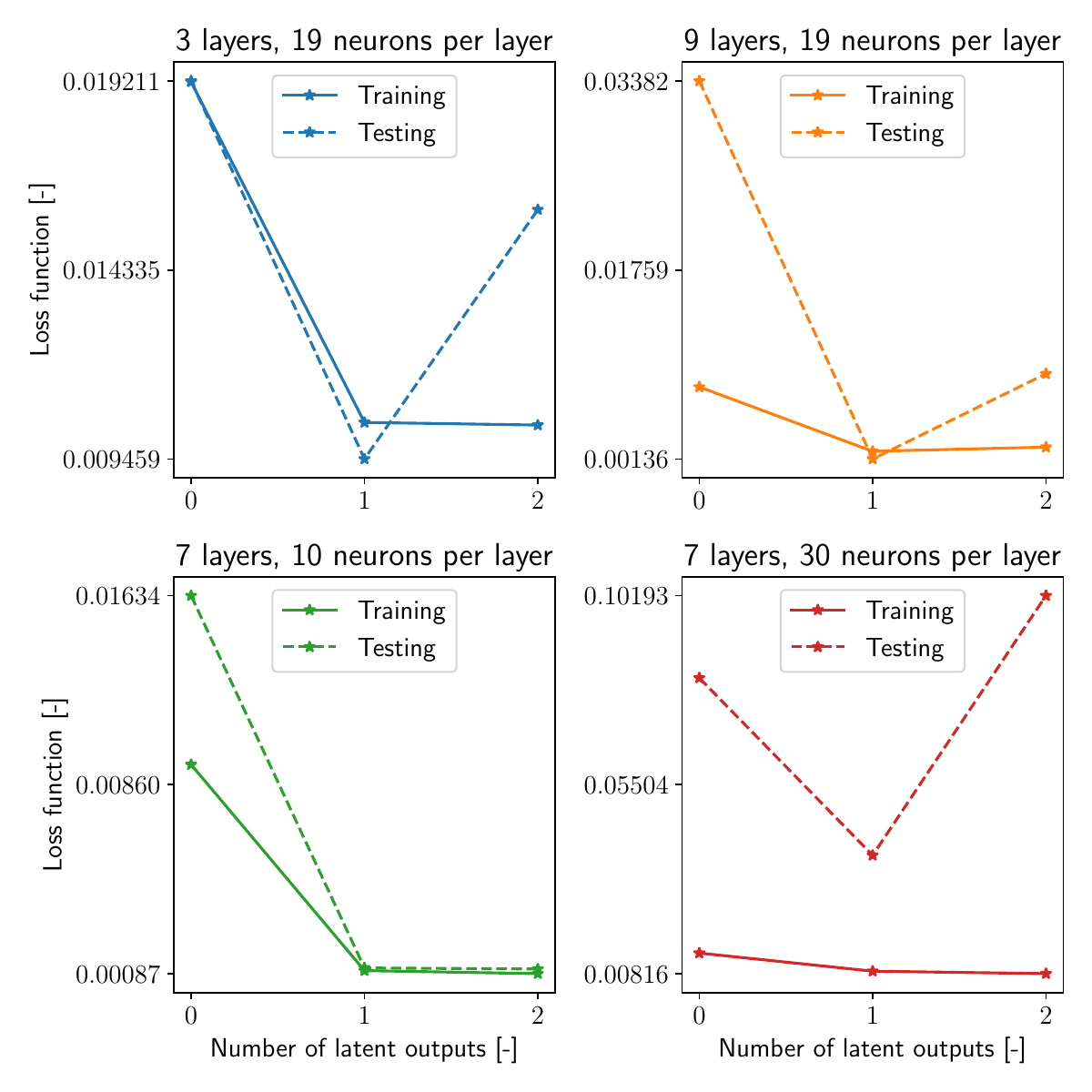}
    \caption{\MSblue{Training and testing errors vs. number of latent outputs associated with four different NN architectures. We consider 50,000 BFGS iterations, 150 and 50 electrophysiology simulations for training and testing, respectively, with $\Delta t = 5.0$ ms.}}
    \label{fig:losslatent}
\end{figure}

We showcase the features of \MS{BLNMs} by means of different test cases.
In Table~\ref{tab:trainingsize} we analyze the influence of the training set size on the computational times and MSE.
We consider the optimal NN architecture obtained from hyperparameters tuning (see Section~\ref{sec:results:tuning}).
We notice that the total training time scales linearly with the dimensionality of the dataset.
Moreover, the training costs on a single CPU are quite modest, approximately ranging from 1 to 3 hours.
The training MSE is small, on the order of $10^{-4}$, and comparable, regardless of the number of electrophysiology simulations.
On the other hand, the testing loss drops to $6 \cdot 10^{-4}$ with 100 numerical simulations.
Given the sharp temporal dynamics of 12-lead ECGs, the number and ranges of model parameters covered by model $\modFOM$, \MS{BLNMs provide} excellent \MS{in-distribution} generalization properties with a relatively small amount of training data, especially when compared to the significant variability and complex dynamics encompassed by the dataset.

In Table~\ref{tab:testingtimestep} \MS{and Figure~\ref{fig:BLNOtestdt}} we study the effect of different testing time steps on the \MS{BLNM} prediction accuracy.
We see that the MSE remains approximately the same on finer and coarser meshes with respect to the fixed time step used for training, that is $\Delta t = 5.0$ ms.
This means that \MS{BLNMs} appear to show little sensitivity to time discretization \MS{even if the training stage is performed on a specific finite dimensional representation of the encoded physical process}.

In Figure~\ref{fig:BLNOtestphysical} we compare the \MS{BLNM} predictions with the ground truth for 5 randomly selected testing samples.
\MS{BLNMs} manifest good agreement with observations, even in presence of sharp peaks and gradients during the QRS complex and T wave propagation.
In Table~\ref{tab:testinglatent} we see the impact of varying the total number of states on the resulting MSEs.
\MS{Adding latent outputs to the 9 physical outputs representing precordial and limb lead recordings allows us to significantly reduce both training and testing errors. Specifically, the training error is approximately halved, whereas the testing error is reduced by two orders of magnitude.}
\MS{This means that the dynamics of 12-lead ECGs can be reproduced more accurately in the presence of a suitable number of latent variables.}
\MS{In particular, from Figure~\ref{fig:BLNOtestlatent} we notice that the additional latent variable selected by the hyperparameter tuning process enhances the BLNM learned dynamics by selectively acting on the QRS complex, that is ventricular depolarization, and T wave, that is ventricular repolarization.}
\MSblue{Similar considerations hold even for sub-optimal NN architectures.}
\MSblue{In Figure~\ref{fig:losslatent} we depict the training and testing errors with respect to the number of latent outputs by considering four different BLNMs with smaller/higher number of layers and/or neurons per layer than the optimal set of hyperparameters. We see that adding one latent output always entails a significant reduction in both loss functions. On the other hand, two latent outputs contribute to a small reduction of the training error while sometimes leading to overfitting. This means that a single latent output is sufficient to capture the required additional features for this specific application.}

\MS{We also train a standard feedforward fully-connected NN with 9 physical outputs, i.e. without latent outputs, 7 layers and 19 neurons per layer, that is the optimal configuration for BLNMs.}
\MS{This NN accounts for 2,631 trainable parameters.}
\MS{We employ the BFGS optimizer and we perform 50,000 epochs over the usual 150 electrophysiology simulations, sampled with $\Delta t = 5.0$ ms.}
\MS{The training error is $7 \cdot 10^{-3}$ while the testing error is $3.1$.}
\MS{This shows that BLNMs outperform standard NNs in terms of training and testing errors while considering less tunable parameters and shorter training times.}
\MS{Moreover, the standard NN does not generalize well on different discretization due to the high MSE reported for the testing loss.}

\MS{Furthermore, we quantitatively compare BLNMs against latent neural differential equations \cite{regazzoni2019modellearning,Chen2019}.}
\MS{We perform hyperparameter tuning using the same ranges reported in Table~\ref{tab:hyperparameters}, except for the disentanglement level, which is not present given the feedforward fully-connected structure of the NN in this framework.}
\MS{Following the approach of BLNMs, we employ the BFGS optimizer and we perform 10,000 epochs, sampling the training and validation sets with a fixed time step $\Delta t = 5.0$ ms.}
\MS{We discretize the latent neural differential equations in time using the forward Euler method, by considering $\Delta t = 5.0$ ms.}
\MS{The optimal configuration of hyperparameters found during $K$-fold ($K = 5$), which is given by 7 layers, 26 neurons per layer and 9 states (i.e no latent variables), has a validation loss that is equal to $54.3$.}
\MS{Indeed, we notice that latent neural differential equations fail to capture the QRS complex and the T wave, which are the most important features of 12-lead ECGs.}

\subsection{\MS{Parameter estimation}}
\label{sec:results:PE}

\begin{figure}[t!]
    \centering
    \includegraphics[width=0.8\textwidth]{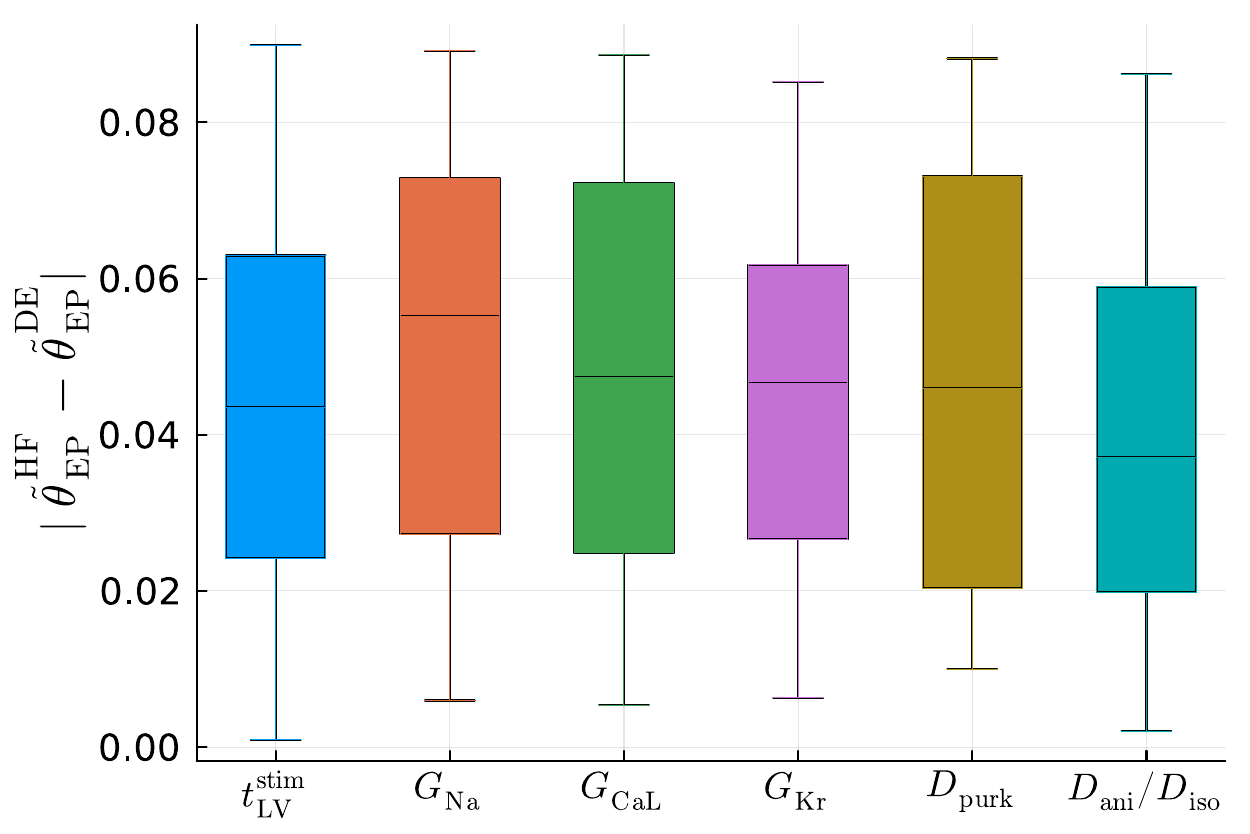}
    \caption{\MS{Box plots showing the distribution of the errors for all model parameters.}}
    \label{fig:BLNOboxplots}
\end{figure}

\MS{We employ the final BLNM to perform parameter calibration against the testing set, which is comprised of 50 electrophysiology simulations.}
\MS{In Figure~\ref{fig:BLNOboxplots} we report the box plots showing the distribution of the errors, given by the absolute difference between each parameter $\widetilde{\theta}_\mathrm{EP}^\mathrm{HF}$ from model $\modFOM$ and each estimated parameter $\widetilde{\theta}_\mathrm{EP}^\mathrm{DE}$ with model $\modROM$, in non-dimensional form.}
\MS{We notice that all the errors are small and lay within the $[0, 0.09]$ range.}
\MS{This is possible given the small approximation error ($\sim$ $10^{-4}$) provided by the BLNM with respect to the high-fidelity electrophysiology simulations.}
\MS{We show that BLNMs can be used to match unseen observations coming from model $\modFOM$, while also retrieving all 7 cell-to-organ model parameters.}
\MS{Performing a single instance of global optimization requires 7 seconds of computations in serial execution on an Intel(R) Core(TM) i7-8700 3.20GHz CPU.}

%% file: parts_discussion.tex
\section{Discussion}
\label{sec:discussion}

Many efforts in the Scientific Machine Learning community are devoted to learning or mapping physical processes, within a certain range of variability, by means of NNs.
This can be performed by either learning the time \cite{regazzoni2019modellearning,Chen2019,Rubanova2019,Dupont2019,Kidger2020}, space \cite{Regazzoni2022USM,Pichi2023} and space-time \cite{Hasani2020,Regazzoni2023,Vlachas2022,Kicica2023,Pegolotti2023} dynamics via different forms of neural differential equations or by mapping the whole solution manifold with physics-informed or data-driven neural \MS{maps} \cite{Kovachki2023,Lu2021,Raissi2019,Raonic2023}.
These involve the use of feedforward fully-connected, recurrent, convolutional or graph neural networks, as well as encoders and decoders based on these architectures.

\MS{BLNMs} blend and share mathematical properties coming from both classes of numerical methods.
Indeed, this novel neural \MS{map encodes} the whole output of interest by spanning model variability in a supervised fashion, while structurally disentangling inputs of different nature, such as time and model parameters of a differential equation.
The level of separation between different categories of inputs can be properly tuned, ranging from the first hidden layer to the outputs of a feedforward partially-connected NN.
\MS{BLNMs} are simple, lightweight architectures, easy and fast to train, that effectively reproduce challenging processes with sharp gradients and fast dynamics in complex solution manifolds.

\MSblue{While autoencoders generally exploit latent variables between the encoder and the decoder in order to perform dimensionality reduction \cite{Fresca2022,Romor2023,SoleraRico2023}, BLNMs are endowed with additional latent outputs that act in specific regions of the simulated process to locally enhance the learned dynamics.}
\MS{This principle is similar to what is done in latent/augmented neural differential equations \cite{Dupont2019,regazzoni2019modellearning,Regazzoni2023}, where the NN defines a novel set of differential equations encoding the dynamics of a specific state vector, which contains both physical and latent variables.}
\MS{The latter are generally not considered in the loss function, as in BLNMs, but allow one to find better dynamics for the physical variables that are targeted during the optimization process.}
\MSblue{BLNMs exploit these latent variables as a lifting in the output dimension in order to better map the whole solution manifold directly, without passing them from a system of differential equations.}
\MS{This enables faster training than neural differential equations, as we do not have to replicate the NN structure over different time steps and we do not need to compute gradients over a chain of NNs during backpropagation.}
\MS{BLNMs require backpropagation over a single NN, where the presence of partial connections significantly reduces the number of tunable parameters with respect to latent neural differential equations.}
\MS{Similar considerations hold for the online inference process, which can be carried out with BLNMs by simply querying the NN without solving one or multiple differential equations.}
\MS{This provides a speed-up in the testing phase of BLNMs in comparison to neural differential equations.}
\MS{Moreover, latent neural differential equations generally struggle to reproduce sharp and irregular features, as we showed in Section~\ref{sec:results:BLNO}.}

\MS{On the other hand, while recent computational tools based on neural differential equations or deep neural operators enable space-time extrapolation \cite{Regazzoni2023,Vlachas2022,Fatone2022,Zhu2023}, learning the input-output map via \MS{BLNMs} currently allows for excellent in-distribution generalization only.}
\MS{Indeed, while testing BLNMs for out-of-distribution generalization, i.e. by considering model parameters outside the training range and longer simulation times, we notice that they provide reasonable approximations only in the neighborhood that is right outside the training range and fail to perform time extrapolation.}
\MS{In particular, after the maximum training time, BLNMs provide the trivial zero solution followed by a divergent behavior.}
\MS{Future studies should aim to improve the performance of BLNMs for out-of-distribution generalization.}

\MS{Another important feature of BLNMs is that they present a comparable performance among different discretizations during the testing phase.}
\MS{This last property is also shared by neural operators \cite{Kovachki2023}, which learn maps between infinite dimensional function spaces.}
\MS{Nevertheless, BLNMs focus on a specific finite dimensional grid during the training stage and are able to generalize over different resolutions during the testing phase.}
\MS{Moreover, if compared to BLNMs, neural operators of different categories, such as Fourier, low-rank, graph-based or deep operators, necessitate a more complex structure within the layers of the NN, which increases training and testing times \cite{Li2021,Lu2021}.}

\MS{BLNMs present several differences with respect to both physics-informed neural networks (PINNs) \cite{Raissi2019} and associated recent extensions \cite{Cuomo2022, Manikkan2023, Pang2020, Penwarden2023}.}
\MS{While both BLNMs and PINNs share a data-driven term in the loss function, the former method encodes latent outputs that enhance the learned dynamics but does not enforce any physics-based knowledge, while the latter focuses on physical outputs only but also incorporates a physics-driven loss function based on the strong form of differential equations.}
\MS{BLNMs focus on a specific mesh during training and present similar generalization errors over both coarser and finer grids during testing.} \MS{On the other hand, PINNs are mesh-less and require a suitable distribution of training points in the parameter space in order to generalize well during the testing phase.}
\MS{BLNMs present a partially connected structure, that allows for reduced complexity (i.e. number of trainable parameters) while structuring separating flows of information coming from inputs that are intrinsically different, whereas PINNs are normally based on fully-connected NNs.}
\MS{Both methods can potentially handle different sets of inputs and outputs, such as space and time variables, scalar and vector fields from parameterized differential equations, model-based or geometrically-based parameters.}
\MS{Furthermore, in this specific application for cardiac electrophysiology, the 12-lead ECGs are obtained by a space integral over the gradient of the transmembrane potential coming from the monodomain equation (see Equation~\eqref{eqn:ECGs}), which makes a direct use of PINNs unfeasible because the physics-based part cannot be incorporated as the residual of a differential equation written in strong form.
On the other hand, BLNMs can properly handle scenarios for model discovery or when the mathematical formulation cannot be seamlessly enforced in the loss function.}

\MS{All the aforementioned aspects characterizing BLNMs are demonstrated on a challenging real-world application in the field of cardiac modeling.}
Specifically, a reduced-order model of in silico 12-lead ECGs spanning 7 cell-to-organ model parameters is learned from biophysically detailed and anatomically accurate electrophysiology simulations on a patient-specific heart-torso geometry of a pediatric patient with HLHS, a complex form of congenital heart disease.
\MS{BLNMs accurately reproduce the outputs of this high-fidelity electrophysiology model and can be readily employed in many-query applications, such as robust and global parameter estimation.}

%% file: parts_conclusions.tex
\section{Conclusions}
\label{sec:conclusions}

We introduced \MS{BLNMs}, a novel computational tool for arbitrary functional mapping.
\MS{BLNMs} structurally disentangles inputs with different intrinsic roles, such as time and model parameters, by means of feedforward partially-connected NNs.
These partial connections can be propagated from the first hidden layer throughout the outputs according to the chosen disentanglement level.
Furthermore, \MS{BLNMs} may be endowed with latent variables in the output space, which enhance the learned dynamics of the neural \MS{map}.

\MS{The novelties of this work reside both in the methods and their application to congenital heart disease, which is understudied in the field of computational cardiology.}
\MS{Indeed,} we apply \MS{BLNMs} in a challenging test case, that is learning the 12-lead ECGs of a pediatric patient with HLHS by covering a large range of 7 significant cell-to-organ model parameters.
We demonstrate that \MS{BLNMs} retain a small number of tunable parameters while accurately encoding complex, irregular and highly variable dynamics.
Moreover, thanks to the efficient Julia implementation, leveraging different NN libraries and optimization tools, these neural \MS{maps} can be trained in a fast manner even on a single CPU.
\MS{BLNMs require} small training datasets and do not degrade in accuracy when tested on a different discretization than the one used for training.
\MS{Furthermore, they can be effectively employed for parameter estimation, as demonstrated using the whole testing set of the high-fidelity numerical simulations.}
\MS{This parameter calibration process can be carried out within a few seconds, i.e. almost in real-time, on a single core standard computer, by considering global optimization in the parameter space.}
\MS{In future works, we aim at using BLNMs to match patient-specific data with numerical simulations by also leveraging computational tools from global sensitivity analysis and robust parameter estimation with uncertainty quantification.}
\MS{Moreover, we would incorporate geometrical features within BLNMs, so that we can cover anatomical variability and we do not need to re-train the NN on every new patient.}

Finally, although we showcased and tested \MS{BLNMs} in a specific application involving time processes only, this paper paves the way for several extensions of the presented approach to space-time processes, while also structurally disentangling different sets of parameters, such as the ones describing geometric variability from scalar and vector values related to a single geometry.
\MS{Furthermore, integrating a physics-based loss or a multifidelity approach, as recently proposed in the framework of deep operator networks \cite{Howard2023}, may improve the performance and generalization of BLNMs, especially for multiscale and multiphysics problems with known physical laws and properties.}

%% file: acknowledgements.tex
\section*{Acknowledgements}

This project has been funded by the NSF SSI grant 1663671, CDSE grant 2105345 and NIH grants R01EB029362, R01LM013120.
We acknowledge Additional Ventures Foundation, Stanford Cardiovascular Institute and the Vera Moulton Wall Center for pulmonary vascular disease at Stanford University.
We thank Dr. Fanwei Kong for the segmentation and mesh generation of the patient-specific heart-torso model.